\documentclass[letterpaper, 10 pt, conference]{ieeeconf}  

\IEEEoverridecommandlockouts                              

\usepackage{multirow}
\usepackage{graphicx}
\usepackage{array}
\usepackage{booktabs}
\usepackage{stmaryrd}
\usepackage{multirow} 
\usepackage{color}
\usepackage{amsmath}
\usepackage{arydshln}
\usepackage{makecell}
\usepackage{pifont}
\usepackage{tabularx}

\newcommand{\redxmark}{\textcolor{red}{\ding{55}}}
\newcommand{\bluecheckmark}{\textcolor{blue}{\ding{51}}}
\usepackage[table]{xcolor} 
\usepackage{tikz}
\definecolor{lightblue}{rgb}{0.8, 0.9, 1.0} 

\usepackage{amsmath, amsfonts, amsthm, stackrel, amssymb}
\usepackage{float}
\usepackage{stfloats}
\usepackage{colortbl}
\usepackage{hyperref}

\title{\LARGE \bf
3D-MoRe: Unified Modal-Contextual Reasoning for Embodied Question Answering
}

\author{Rongtao Xu$^{*}$, Han Gao$^{*}$, Mingming Yu, Dong An, Shunpeng Chen, Changwei Wang, \\ Li Guo, Xiaodan Liang$^{\dag}$, Shibiao Xu}

\begin{document}

\maketitle

\renewcommand{\thefootnote}{}
\footnotetext{
$^{\dag}$Xiaodan Liang is the corresponding author (xdliang328@gmail.com).
$^{*}$Rongtao Xu and Han Gao contributed equally.
Rongtao Xu is with Spatialtemporal AI. Han Gao, Shunpeng Chen, Li Guo and Shibiao Xu is with the Beijing University of Posts and Telecommunications, China. Mingming Yu and Dong An are with the Institute of Automation, Chinese Academy of Sciences, China. Changwei Wang are with the Key Laboratory of Computing Power Network and Information Security, Ministry of Education; Shandong Computer Science Center. Xiaodan Liang is with Sun Yat‑Sen University, China.

}

\thispagestyle{empty}
\pagestyle{empty}

\begin{abstract}

With the growing need for diverse and scalable data in indoor scene tasks, such as question answering and dense captioning, we propose 3D-MoRe, a novel paradigm designed to generate large-scale 3D-language datasets by leveraging the strengths of foundational models. The framework integrates key components, including multi-modal embedding, cross-modal interaction, and a language model decoder, to process natural language instructions and 3D scene data. This approach facilitates enhanced reasoning and response generation in complex 3D environments. Using the ScanNet 3D scene dataset, along with text annotations from ScanQA and ScanRefer, 3D-MoRe generates 62,000 question-answer (QA) pairs and 73,000 object descriptions across 1,513 scenes. We also employ various data augmentation techniques and implement semantic filtering to ensure high-quality data. Experiments on ScanQA demonstrate that 3D-MoRe significantly outperforms state-of-the-art baselines, with the CIDEr score improving by 2.15\%. Similarly, on ScanRefer, our approach achieves a notable increase in CIDEr@0.5 by 1.84\%, highlighting its effectiveness in both tasks. Our code and generated datasets will be publicly released to benefit the community, and both can be accessed on the \href{https://3d-more.github.io/}{\textcolor{magenta}{https://3D-MoRe.github.io}}.

\end{abstract}


\section{INTRODUCTION}

\begin{figure}[!t]
\centering
\includegraphics[width=0.9\linewidth]{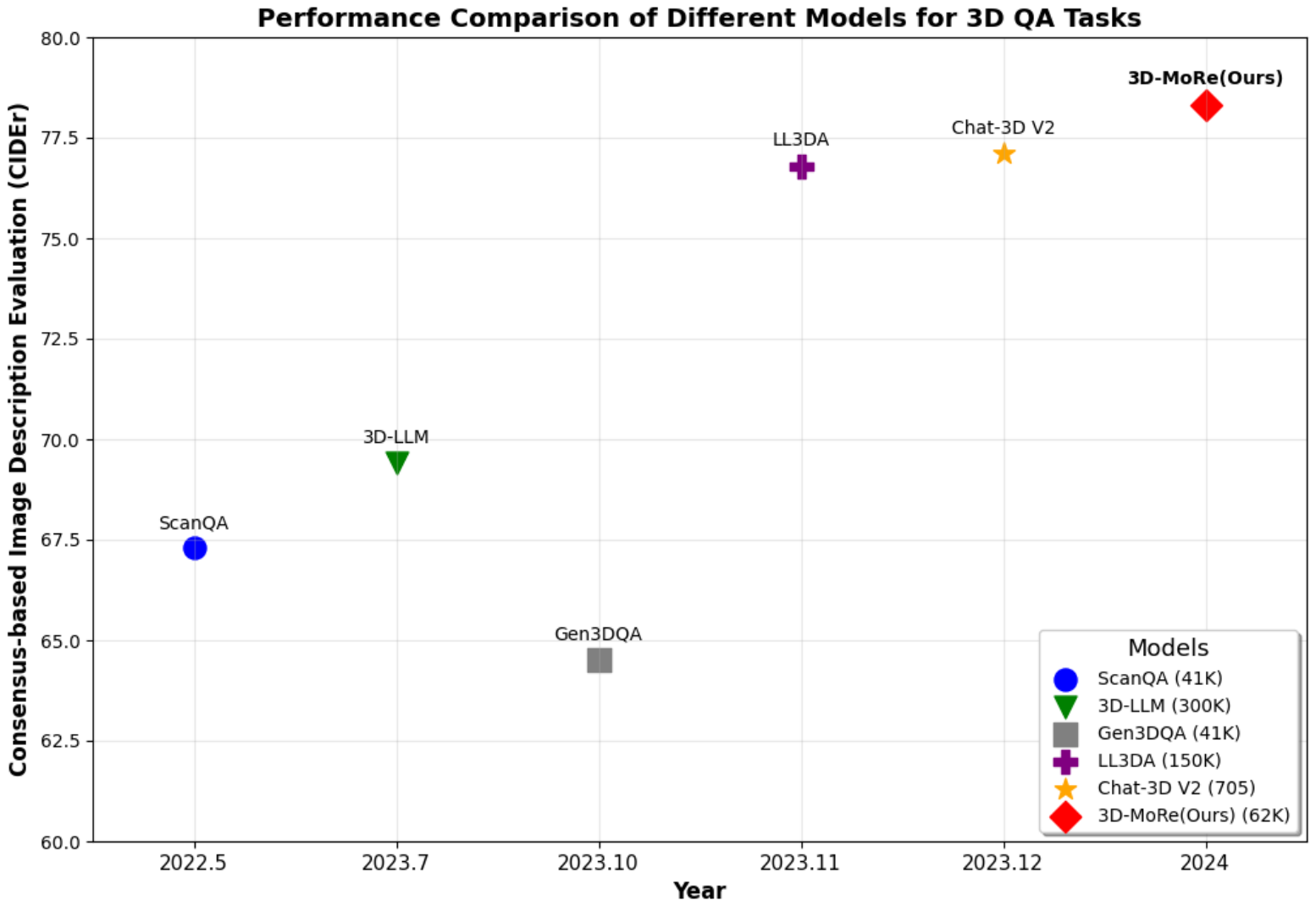}
\vspace{-0.2cm}
\caption{The agent's success rate in 3D Question Answering improves with increased data size. Our method generates 62K QA pairs, substantially enhancing performance and bringing it closer to human-level results.}
\label{3dqa_CIDEr}
\vspace{-0.5cm}
\end{figure}

In 3D question answering (3DQA) and dense captioning tasks, models must tackle complex multimodal reasoning within 3D environments. The 3DQA task demands deep scene understanding and spatial reasoning to answer text-based questions, while dense captioning requires detailed descriptions of objects and their relationships in 3D space. Existing methods often rely on multimodal fusion techniques such as semantic-level data augmentation~\cite{xu2023scd,xu2023rssformer,xu2023wave}, spatial attention~\cite{zhang2025robridge}, and cross-modal encoding to enhance object localization and scene comprehension~\cite{xu2024local,xu2023domainfeat,wang2025focus}. Unlike Gen3DQA~\cite{dwedari2023generating}, which focuses on small datasets, our 3D-MoRe method leverages diverse data augmentation strategies to broaden dataset variety. In contrast to Vote2Cap-DETR++~\cite{chen2024vote2cap}, which depends heavily on spatial features, our approach integrates both spatial and linguistic information to achieve robust performance across various environments. Advanced semantic filtering ensures high-quality data, significantly improving contextual accuracy.

Generating large-scale datasets poses challenges, including prompt construction\cite{white2023prompt,yan2024instrugen,zhang2024navid}, accurate annotation extraction~\cite{falcon2022feature}, and data quality filtering~\cite{vadlapati2024autopuredata}. To address these, we propose the Adaptive Multimodal Fusion Paradigm, which incorporates three generation methods: QA Generation (expanding ScanQA), Captioning Generation (transforming ScanRefer captions into QA pairs), and Scene Generation (using vision-language models to generate QA pairs from 3D scene data). Additionally, we introduce two data filtering techniques—semantic similarity and search—to ensure data quality. This approach generates 62,000 triplets for the 3DQA task and 73,000 for captioning, as shown in Figure~\ref{3dqa_CIDEr}. Using these triplets, we train a 3D-LLM model that encodes the triplets across three branches, aligns the modalities through interaction, and decodes responses with an LLM, achieving a significant performance improvement, approaching human-level proficiency.

In summary, our key contributions lie in:

\begin{itemize}
\item We introduce 3D-MoRe, an innovative framework that leverages foundational models to generate large-scale 3D-language datasets, integrating multimodal embedding, cross-modal interaction, and a language model decoder to enhance reasoning in complex 3D environments.

\item 3D-MoRe synthesizes 62,000 question-answer pairs and 73,000 object descriptions from the ScanNet dataset, significantly increasing data diversity and improving performance on 3D question answering and 3D dense captioning tasks.

\item Through advanced data augmentation techniques, including synonym substitution, sentence reordering, and semantic filtering, 3D-MoRe achieves a 2.15\% improvement in CIDEr on ScanQA and a 1.84\% increase in CIDEr@0.5 on ScanRefer, effectively enhancing model accuracy.

\end{itemize}

\begin{figure*}[htbp]
\centering
\includegraphics[width=0.93\linewidth]{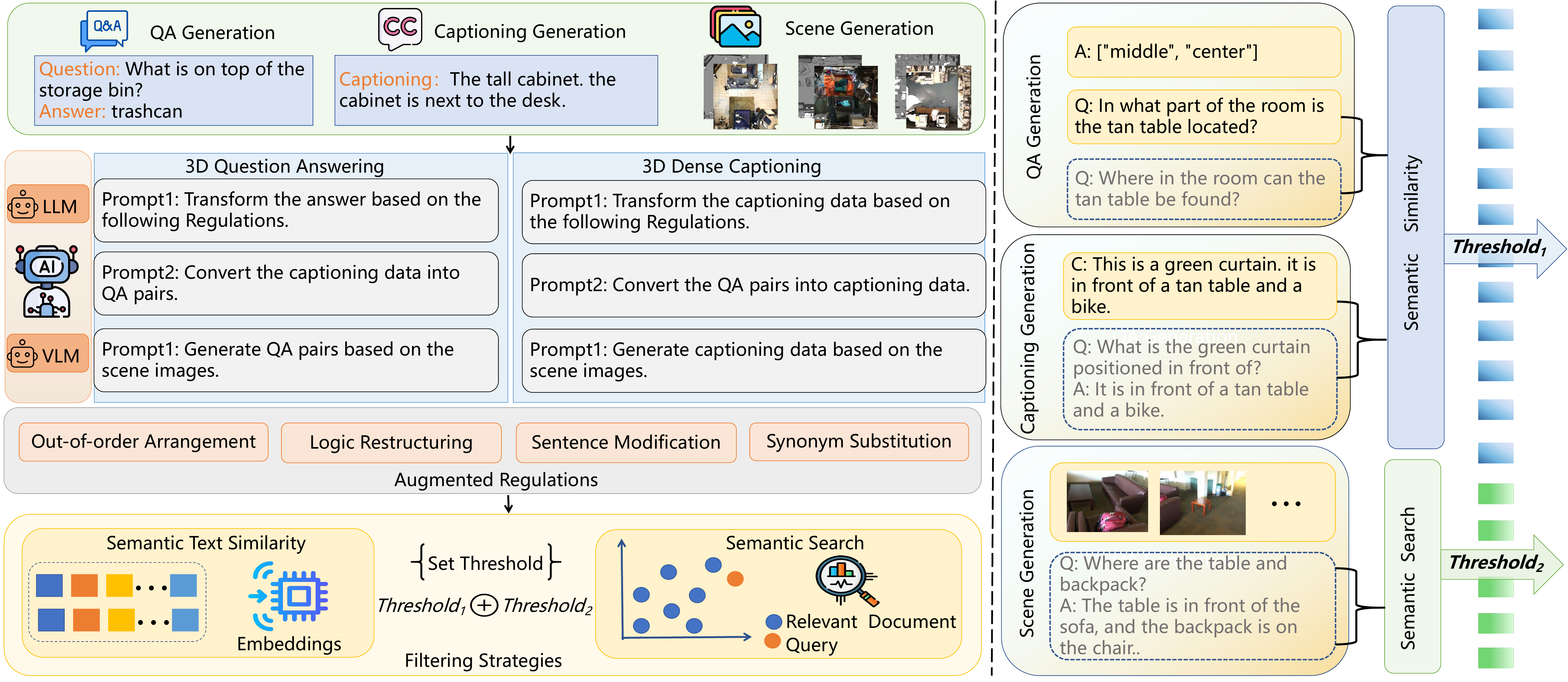}
\vspace{-0.2cm}
\caption{\textbf{Generation paradigm pipeline.} To expand the datasets, we combine ScanNet scene data with textual annotations from ScanQA and ScanRefer. We apply semantic search and similarity filtering to rigorously select generated data and obtain high-quality text embeddings. The figure on the right illustrates the filtering strategies for the 3D Question Answering task, with parentheses indicating the compared text data.}
\label{data_generation}
\vspace{-0.7cm}
\end{figure*}

\section{Related Work}

\textbf{3D Question Answering and 3D Dense Captioning.}  
3D Question Answering and Dense Captioning involve interpreting 3D scenes by utilizing depth and point cloud data to enhance spatial understanding. 3DQA models align visual and linguistic data to improve response accuracy and reduce uncertainty~\cite{ma2024robust}. Datasets like ScanQA~\cite{azuma2022scanqa}, CLEVR3D~\cite{yan2023comprehensive}, and FE-3DGQA~\cite{zhao2022toward} provide essential benchmarks. Dense Captioning generates detailed descriptions of objects and their spatial relationships, leveraging depth information for better object geometry capture~\cite{yu2023comprehensive}\cite{chen2023unit3d}. Transformer architectures combined with point cloud networks enhance the alignment of 3D visual features with language representations~\cite{yuan2022x}. Datasets like ScanRefer and ScanNet support model training with rich 3D annotations~\cite{chen2020scanrefer}\cite{dai2017scannet}.

\textbf{Large Language Models.} 
Recent advancements in large language models (LLMs) have enabled complex reasoning and conversational understanding, fueled by internet-scale data~\cite{xu2025a0,han2025multimodal,zhang2025activevln,ma2025phyblock}. Recent work extends LLM capabilities to visual reasoning tasks, advancing multimodal processing~\cite{chung2024scaling}\cite{han2025multimodal}. Vision-language models like LLaVA~\cite{liu2024visual} use LLMs to generate question-answer pairs from image descriptions, improving performance in 2D tasks. However, research on 3D scene instruction-tuning remains limited. Our work enhances model capabilities by generating triplets of 3D point clouds, visual prompts, and text instructions to improve 3D understanding and reasoning.

\section{Methodology}
\subsection{Problem Formulation}

We aim to train a generalist agent capable of handling various 3D-language tasks using samples from our proposed scaling data paradigm. The agent processes a 3D scene context represented as point clouds, visual prompts such as 3D bounding boxes and instance prompts, and natural language instructions. It needs to understand both the textual instructions and the 3D scene, interpreting spatial and contextual information to generate an appropriate natural language response.

\subsection{Adaptive Multimodal Fusion Paradigm}
\label{scaling_data_paradigm}

As formalized in Figure.~\ref{data_generation}, Our framework implements quality-controlled data augmentation through metric-guided transformations. By integrating multi-source inputs $\mathcal{D} = \{D_{\text{scene}}, D_{\text{QA}}, D_{\text{cap}}\}$ from ScanNet \cite{dai2017scannet}, ScanQA \cite{azuma2022scanqa}, and ScanRefer \cite{chen2020scanrefer}, the augmentation pipeline applies task-specific transformations $\Phi_k$ followed by metric-based filtering $\Psi_k$, yielding the final dataset $\mathcal{D}_{\text{final}} = \bigcup_{k=1}^3 \Phi_k \circ \Psi_k(D_k)$.

\subsubsection{Semantic Quality Control}
To ensure high-quality data generation, we rely on two key metrics. First, semantic similarity is measured as $S_Q(Q_{\text{orig}}, Q_{\text{gen}}) = \cos\bigl(f_{\text{BERT}}(Q_{\text{orig}}), f_{\text{BERT}}(Q_{\text{gen}})\bigr)$, where BERT embeddings \cite{devlin2018bert} quantify the alignment between the original and generated questions. Second, we assess semantic consistency via a semantic search approach, defined as $S_{\text{cap}}(C_{\text{orig}}, C_{\text{gen}}) = \frac{1}{n} \sum_{i=1}^n \text{NLI}(C_{\text{orig}}, C_{\text{gen}}^{(i)})$, which leverages RoBERTa inference \cite{yinhan2019roberta} to evaluate the correctness of caption-derived QA pairs. Task-specific thresholds, determined from human-annotated statistics as $\tau_k = \mu_k + 1.96\sigma_k$, are set to $\tau_{\text{QA}}=0.82$ for QA tasks and $\tau_{\text{cap}}=0.77$ for captioning.

\subsubsection{Augmentation Architectures}
For QA generation, we apply transformations such as synonym replacement, logical reversal, and order shuffling, with relevance scoring computed as $\text{rel}(A,Q) = \sigma(\mathbf{W}_r[f_{\text{BERT}}(A); f_{\text{BERT}}(Q)])$, where $\sigma$ denotes the sigmoid function. Caption-to-QA conversion employs a T5 model \cite{pilehvar2018wic}, where the generated question $Q_{\text{gen}} = \text{T5}(C \oplus \mathcal{P}_{\text{template}})$ is produced using 32 handcrafted templates $\mathcal{P}_{\text{template}}$, and answers are projected via $\mathbf{W}_{\text{ans}} \in \mathbb{R}^{d \times |\mathcal{V}|}$.

\subsubsection{Multimodal Integration}
For scene-to-QA generation, we adopt CogVLM \cite{hong2024cogagent}, where QA likelihood is computed through 3D-text cross-attention as $p(Q,A \mid S) = \text{softmax}(\text{CrossAtt}(\mathbf{E}_{\text{3D}}, \mathbf{E}_{\text{text}}))$. The final dataset $\mathcal{D}_{\text{final}}$ integrates over 62,000 QA pairs and 73,000 caption
annotations, surpassing the original ScanQA and ScanRefer datasets.

\begin{figure*}[htbp]
    \centering
    \includegraphics[width=1\linewidth]{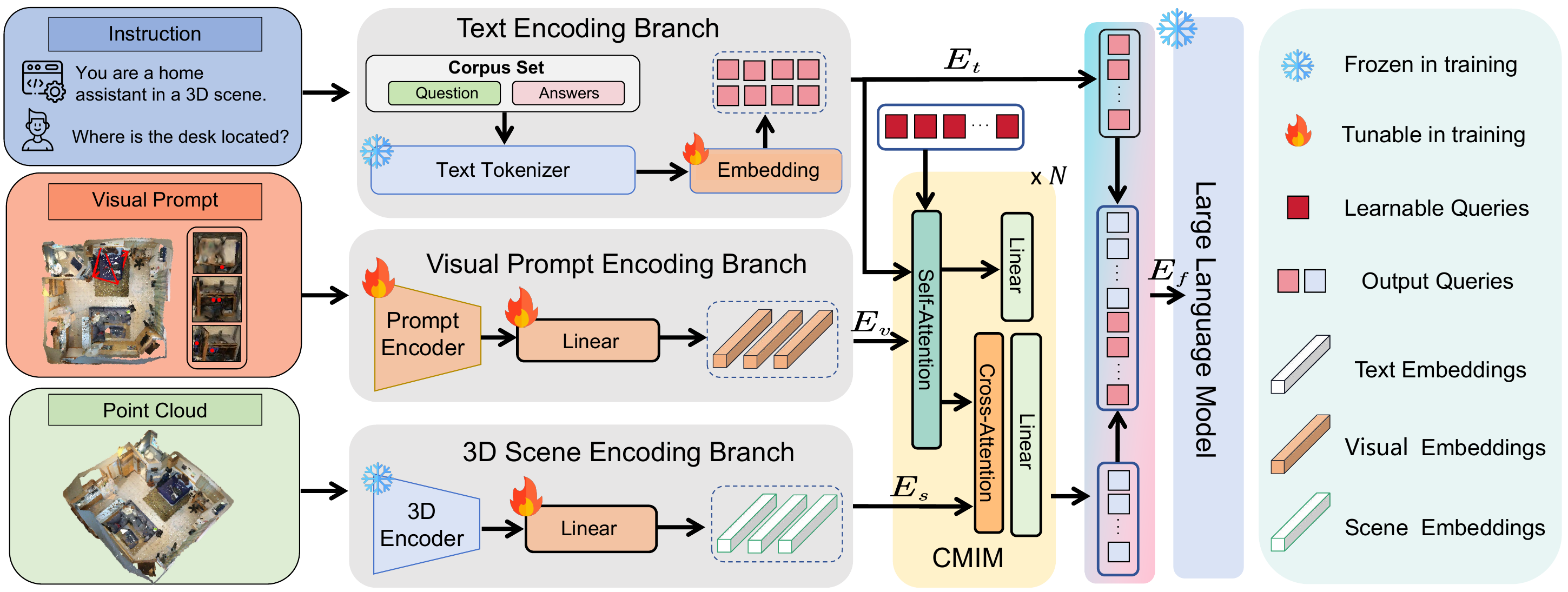}
    \vspace{-0.2cm}
    \caption{The architecture processes multi-modal inputs, combining natural language and 3D scene data for reasoning in 3D environments. It features three main components: Multi-Modal Embedding, Cross-Modal Interaction, and LLM Decoder. \textbf{CMIM}: Cross-Modal Interaction Module.}.
\label{model_diagram}
\vspace{-0.2cm}
\end{figure*}

\subsection{Model Architecture}
\label{sec:arch}

The proposed architecture establishes a hierarchical fusion framework for embodied reasoning, comprising three core components: 1) Multi-Modal Embedding Module, 2) Cross-Attention Fusion Module, and 3) LLM Decoder. As formalized in Figure.~\ref{model_diagram}, the processing pipeline $\mathcal{R} = \text{LLM}_{\text{dec}}(\mathcal{F}_{\text{fusion}}(\mathcal{E}_t(T) \parallel \mathcal{E}_v(V) \parallel \mathcal{E}_s(S)))$ integrates text $T$, visual prompts $V$, and 3D scene $S$ through tensor concatenation.

\subsubsection{Multi-Modal Embedding Module}
\begin{itemize}
    \item \textbf{Text Encoding}: For input tokens $T = \{w_i\}_{i=1}^L$, standard Transformer encoding generates embeddings $\mathbf{E}_t = \text{Transformer}_{\text{enc}}(\text{Embed}(T) + \mathbf{P}_t) \in \mathbb{R}^{L \times d}$ with positional encoding $\mathbf{P}_t$.
    
    \item \textbf{Visual Prompt Encoding}: Spatial guidance signals $V = \{v_j\}_{j=1}^{N_v}$ (3D bounding boxes $v_j = (\mathbf{c}_j, \mathbf{d}_j) \in \mathbb{R}^6$) are obtained through: 
    \begin{enumerate}
        \item User annotations  \item Mask3D detector outputs
    \end{enumerate}
    Encoded via $\mathbf{E}_v = \text{MLP}(\text{Flatten}(V)) \in \mathbb{R}^{N_v \times d}$ to establish spatial priors.
    
    \item \textbf{3D Scene Encoding}: Point cloud $S \in \mathbb{R}^{N_p \times 3}$ is processed by Vote2Cap-DETR++:
    \begin{equation}
        \mathbf{E}_s = \text{Vote2Cap-DETR++}(S) \in \mathbb{R}^{N_s \times d}
    \end{equation}
\end{itemize}

\subsubsection{Cross-Attention Fusion Module}
The fusion module implements three-stage feature integration:

1. \textbf{Cross-Modal Alignment}: 
\begin{align}
    \mathbf{E}_{f1} &= \text{softmax}\left(\frac{\mathbf{E}_t\mathbf{W}_q(\text{Concat}(\mathbf{E}_v,\mathbf{E}_s)\mathbf{W}_k^\top}{\sqrt{d}}\right) \cdot (\mathbf{E}_v \parallel \mathbf{E}_s)\mathbf{W}_v
\end{align}

2. \textbf{Context Preservation}: Self-attention processes $[\mathbf{E}_{f1};\mathbf{E}_t]$ through Transformer layers to maintain linguistic coherence while enabling adaptive fusion control \cite{vaswani2017attention}.

3. \textbf{Residual Fusion}: 
\begin{equation}
    \mathbf{E}_f = \text{LayerNorm}(\text{TransformerLayer}(\mathbf{E}_{f1} \parallel \mathbf{E}_t) + \mathbf{E}_t)
\end{equation}
preserving instructional details via residual connections \cite{he2016deep}.

\subsubsection{LLM Decoder}
The decoder implements visual-grounded generation through dynamic prefix projection:
\begin{equation}
    \mathbf{h}_t = \text{LLM}(\mathbf{p}_{1:t}; \text{LinearProj}(\mathbf{E}_f))
\end{equation}
where visual-spatial context flows via projected prefix embeddings to condition token probabilities $p(w_{t+1}) = \text{softmax}(\mathbf{h}_t \mathbf{W}_{\text{vocab}})$.

\section{Experiments}

\subsection{Implementation Details}
We trained our model on the ScanQA~\cite{azuma2022scanqa} dataset and object descriptions from ScanRefer~\cite{chen2020scanrefer}, extending the data with an additional 36,437 QA pairs and 36,635 captioning instances. Generated textual responses were evaluated using BLEU~\cite{papinesi2002bleu}, ROUGE~\cite{chin2004rouge}, METEOR~\cite{banerjee2005meteor}, and CIDEr~\cite{vedantam2015cider} metrics. The LL3DA framework~\cite{chen2024ll3da} was employed, with 40,000 points randomly sampled from each 3D scene as input. The model used the frozen OPT-1.3B~\cite{zhang2022opt} as the LLM decoder and Vote2Cap-DETR~\cite{chen2023end} as the object detector in the visual prompt encoder. Training was conducted over 10,000 iterations using the AdamW optimizer~\cite{loshchilov2017decoupled}, with a cosine annealing learning rate schedule, on two Nvidia RTX 3090 GPUs with a batch size of 24.

\begin{table*}[htbp]
\vspace{-0.7cm}
\centering
\caption{Comparison of results on ScanQA for related work. \textbf{multimodal combination:} The generated data comes from ScanQA, ScanRefer text annotation data, and Scannet scene data. \textbf{Params}: Model parameter size.}
\renewcommand{\arraystretch}{1.3}
\begin{tabular}{c|c|c|cccc}
\hline \hline
\multirow{2}{*}{\textbf{Method}}  & \multirow{2}{*}{\textbf{Version}} & \multirow{2}{*}{\textbf{Params}} & \multicolumn{4}{c}{\textbf{Validation}}                                      \\ \cline{4-7} 
                                  &                         &                         & \textbf{CIDEr↑} & \textbf{BLEU-4↑} & \textbf{METEOR↑} & \textbf{ROUGE-L↑}                        \\ \hline \hline
\multirow{3}{*}{3D-LLM\cite{hong20233d}}  & flamingo                & 3B                      & 59.20       & 7.20          & 12.20       & 32.30                              \\
                                  & BLIP2-opt               & 3B                      & 63.80       & 9.40          & 13.80       & 34.00                              \\
                                  & BLIP2-flant5            & 3B                      & 69.40       & 12.00         & 14.50       & 35.70                              \\ \hdashline
\multirow{2}{*}{LL3DA\cite{chen2024ll3da}}   & scratch                & 1.3B                    & 74.80          & 13.68          & 15.40          & \multicolumn{1}{c}{36.25} \\
                                  & fine-tuned              & 1.3B                    & 76.79       & 13.53         & 15.88       & 37.31                              \\ \hdashline
\multirow{3}{*}{GPT-4\cite{singh2024evaluating}}   & GPT Blind             & 1800B                   & 53.59       & 3.81          & 13.54       & 30.92                              \\
                                  & Vocab-agnostic          & 1800B                   & 34.22       & 0.98          & 8.75        & 20.03                              \\
                                  & Vocab-grounded          & 1800B                   & 58.32       & 1.63          & 14.23       & 33.43                              \\ \hdashline
\multirow{2}{*}{Gen3DQA\cite{dwedari2023generating}} & single object        & -                       & 64.91          & 10.52          & 13.62          & 33.39                     \\
                                  & multiple objects        & -                       & 64.51       & 10.21         & 13.68       & 32.84                                                          \\ \hdashline
\multirow{3}{*}{3DMIT\cite{li20243dmit}}   & Vicuna-7b               & -                       & 44.38       & 6.44          & 10.40       & 24.64                              \\
                                  & LLaVA1.5+IMG            & -                       & 46.42       & 5.98          & 10.64       & 24.46                              \\
                                  & Vicuna-7b               & -                       & 48.03       & 5.24          & 10.70       & 26.22                              \\ \hdashline
NaviLLM\cite{zheng2024towards}                     & Vicuna-7b & 7B                    & 75.9 & 12.5 & 15.40 & \multicolumn{1}{c}{38.40} \\ \hdashline
\multirow{2}{*}{Chat-3D\cite{huang2023chat}} & Chat-3D                 & 7B                      & 53.20       & 6.40          & 11.90       & 28.50                              \\
                                  & Chat-3D V2              & 7B                      & 77.10       & 7.30          & \textbf{16.10}       & \multicolumn{1}{c}{\textbf{40.10}} \\ \hline                                  
\textbf{3D-MoRe(Ours)}                     & multimodal combination & 1.3B                    & \textbf{78.94} & \textbf{14.17} & 16.07 & \multicolumn{1}{c}{37.89} \\ 
\hline \hline
\end{tabular}
\label{tab:Comparison of results}
\end{table*}

\begin{table*}[htbp]
\vspace{-0.3cm}
\centering
\small  
\caption{Comparison of results on ScanRefer for related work. \dag:The dataset used comes from multimodal combination.}
\renewcommand{\arraystretch}{1.3}
\begin{tabular}{c|c|cccc}
\hline \hline
\multirow{2}{*}{\textbf{Method}} & \multirow{2}{*}{\textbf{Params}} & \multicolumn{4}{c}{\textbf{ScanRefer}} \\ \cline{3-6} 
 &  & \textbf{CIDEr@0.5↑} & \textbf{BLEU-4@0.5↑} & \textbf{METEOR@0.5↑} & \textbf{ROUGE-L@0.5↑} \\ \hline
Vote2Cap-DETR\cite{chen2024vote2cap} & 60M    & 61.19 & 34.46 & 26.22 & 54.40 \\
3D-VisTA\cite{zhu20233d}             & 0.12B  & 61.60 & 34.10 & \textbf{26.80} & \textbf{55.00} \\
X-Trans2Cap\cite{yuan2022x}          & 0.12B  & 43.87 & 25.05 & 22.46 & 45.28 \\
UniT3D\cite{chen2023unit3d}          & -      & 46.69 & 27.22 & 21.91 & 45.98 \\
3D-VLP\cite{jin2023context}          & -      & 54.94 & 32.31 & 24.83 & 51.51 \\
LL3DA\cite{chen2024ll3da} & 1.3B   & 62.76 & 35.00 & 25.68 & 54.23 \\
MORE\cite{jiao2022more}              & -      & 40.94 & 22.93 & 21.66 & 44.42 \\ \hline
\textbf{3D-MoRe(ours)} & 1.3B   & \textbf{64.08} & \textbf{35.52} & 25.90 & 54.49 \\ \hline \hline
\end{tabular}
\label{tab:Comparison of results2}
\vspace{-0.3cm}
\end{table*}

\subsection{Comparison With Leading Methods}
Table.\ref{tab:Comparison of results} and Table.\ref{tab:Comparison of results2} present the performance of our approach with existing methods. 
Here, "Versions" refer to different model architectures or experimental techniques. Our model consistently outperformed previous methods, especially on the validation set, using the CIDEr metric as the primary indicator. Notably, our approach surpassed the generation-based Chat-3D V2 model~\cite{huang2023chat}, improving the CIDEr@0.5 score by 1.84\%. Additionally, in a comparison on the ScanRefer dataset, our model outperformed LL3DA and other prior approaches, showing a 2.15\% improvement in CIDEr performance on low-parameter LLMs.

\begin{table*}[htbp]
\caption{Comparison of dataset types using 28K samples for 3DQA and 39K for 3D Dense Captioning (Control the comparison quantity,full results in Table~\ref{tab:Comparison of results}). \textbf{Quality Control}: 3DQA uses semantic search; 3D Dense Captioning uses semantic similarity. \textbf{Combination Gen}: A mix of QA, Captioning, and Scene generation methods.}
\centering
\small
\renewcommand{\arraystretch}{1.0}
\setlength{\tabcolsep}{8pt} 
\begin{tabular}{@{}lcccccc@{}}
\toprule
\multirow{2}{*}{Dataset} & 
\multirow{2}{*}{\shortstack{Quality\\Control}} & 
\multicolumn{4}{c}{Core Metrics} & 
\multirow{2}{*}{\shortstack{Data\\Size}} \\
\cmidrule(lr){3-6}
& & 
\shortstack{CIDEr(QA)\\CIDEr@0.5(DC)} & 
\shortstack{BLEU-4(QA)\\BLEU-4@0.5(DC)} & 
\shortstack{METEOR(QA)\\METEOR@0.5\\ (DC)} & 
\shortstack{ROUGE-L(QA)\\ROUGE-L@0.5(DC)} & \\ 
\midrule

\multicolumn{7}{l}{\textbf{3D Question Answering (28K Data)}} \\
\midrule
ScanQA  & - & 76.79 & 13.53 & \textbf{15.88} & 37.31 & 25K \\ 
\multirow{2}{*}{ScanQA + QA Gen} & \redxmark & 77.81 & 13.69 & 15.58 & 37.67 & 28K \\
 & \bluecheckmark & 78.18 & 14.52 & 15.65 & 37.82 & 28K \\ 
\multirow{2}{*}{ScanQA + Captioning Gen} & \redxmark & 77.97 & 14.21 & 15.55 & 37.63 & 28K \\ 
& \bluecheckmark & 78.24 & 14.50 & 15.60 & 37.66 & 28K \\ 
\multirow{2}{*}{ScanQA + Scene Gen} & \redxmark & 77.74 & 14.61 & 15.60 & 37.59 & 28K \\
& \bluecheckmark & 78.32 & 14.59 & 15.73 & 37.78 & 28K \\
ScanQA + Combination Gen & \bluecheckmark & \textbf{78.43} & \textbf{14.62} & 15.75 & \textbf{37.84} & 28K \\
\midrule

\multicolumn{7}{l}{\textbf{3D Dense Captioning (39K Data)}} \\
\midrule
ScanRefer  & - & 62.76 & 35.00 & 25.68 & 54.23 & 36K \\
\multirow{2}{*}{ScanRefer + QA Gen} & \redxmark & 62.71 & 34.89 & 25.41 & 54.12 & 39K \\
& \bluecheckmark & 62.77 & 35.16 & 25.70 & 54.28 & 39K \\
\multirow{2}{*}{ScanRefer + Captioning Gen} & \redxmark & 63.08 & 35.11 & 25.29 & 54.01 & 39K \\
& \bluecheckmark & 63.15 & 35.23 & 25.69 & 54.33 & 39K \\
\multirow{2}{*}{ScanRefer + Scene Gen} & \redxmark & 62.75 & 35.14 & 25.53 & 54.17 & 39K \\
& \bluecheckmark & 62.79 & 35.21 & 25.74 & 54.30 & 39K \\
ScanRefer + Combination Gen & \bluecheckmark & \textbf{63.21} & \textbf{35.33} & \textbf{25.77} & \textbf{54.38} & 39K \\
\bottomrule
\end{tabular}
\label{tab:unified_comparison}
\vspace{-0.4cm}
\end{table*}

\subsection{Ablation Study}

\textbf{Data Augmentation and Filtering Strategies}

In this study, we adopted the Adaptive Multimodal Fusion Paradigm (Sec.~\ref{scaling_data_paradigm}) to expand our datasets for 3D Question Answering and 3D Dense Captioning tasks. To ensure a fair comparison among different data generation methods, we uniformly sampled 3,000 instances from each of the three types—QA Generation (derived from question-answer pairs), Captioning Generation (from dense captions), and Scene Generation (based on image scene data)—implemented via the Sentence Transformers framework. For 3D Question Answering, additional question-answer pairs were generated and subsequently filtered using semantic search to remove low-quality data based on their alignment with the original pairs, whereas for 3D Dense Captioning, semantic similarity measures were employed to select high-quality captions. As demonstrated in Tables~\ref{tab:unified_comparison} , our combined data augmentation and filtering approach achieved the highest CIDEr scores of 78.43\% and 63.21\% for 3D Question Answering and Dense Captioning, respectively, while consistently improving performance across all augmented data. Our code and dataset have been made openly available.

\begin{table}[htbp]
\vspace{-0.2cm}
\caption{The impact of object detection on experimental outcomes.}
\renewcommand{\arraystretch}{1.4} 
\centering
\begin{tabular}{cccccc}
\hline
\multirow{2}{*}{Task} & Object & \multicolumn{4}{c}{Evaluation Criteria} \\ \cline{3-6} 
 & Detection & C\textsuperscript{*} & B-4\textsuperscript{*} & M\textsuperscript{*} & R\textsuperscript{*} \\ \hline
\multirow{2}{*}{\begin{tabular}[c]{@{}c@{}}3D Question\\ Answering\end{tabular}} & \redxmark & 77.58 & 14.11 & 15.85 & 37.70 \\
 & \bluecheckmark & 78.69 & 14.07 & 16.04 & 38.04 \\ \hdashline
\multirow{2}{*}{\begin{tabular}[c]{@{}c@{}}3D Dense\\ Captioning\end{tabular}} & \redxmark & 63.26 & 36.44 & 25.99 & 54.81 \\
 & \bluecheckmark & 63.74 & 36.16 & 26.16 & 54.97 \\ \hline
\end{tabular}
\label{tab:Optimizing Object Detection for Visual Excellence}
\vspace{-0.4cm}
\end{table}

\begin{figure}[ht]
\vspace{-0.2cm}
\centering
\includegraphics[width=0.94\linewidth]{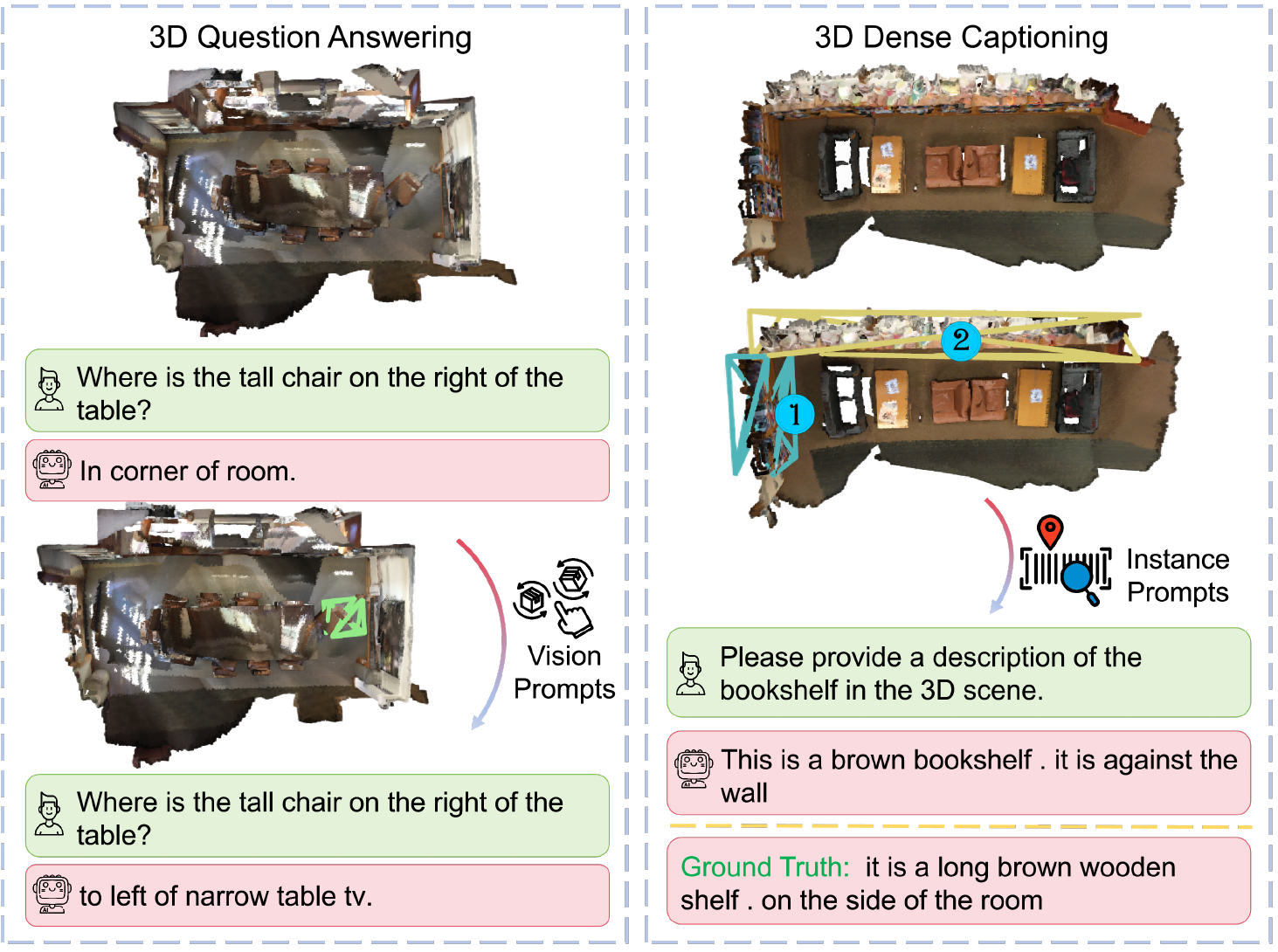}
\caption{Vision and instance prompts enhance object localization and differentiation, improving model accuracy in 3D captioning and question answering.}
\label{qualitative_results}
\vspace{-0.5cm}
\end{figure}

\textbf{Effectiveness of the Visual Prompt.}  
Table~\ref{tab:Optimizing Object Detection for Visual Excellence} shows that incorporating object detection as a visual prompt significantly improves performance. Without it, key metrics in 3D Question Answering (CIDEr, BLEU-4, METEOR, ROUGE-L) and 3D Dense Captioning (CIDEr@0.5, METEOR@0.5, ROUGE-L@0.5) decline noticeably. These results underscore that object detection is vital for capturing spatial context and generating accurate, contextually relevant descriptions, thereby enhancing multimodal reasoning in 3D environments.

\subsection{Qualitative Results}
We evaluated performance in 3D Dense Captioning and Question Answering (Figure~\ref{qualitative_results}) by analyzing cases and incorporating failure analysis with qualitative examples and quantitative error breakdowns (e.g., attribute, spatial, semantic). This revealed spatial reasoning limitations in crowded scenes. Consistency was enhanced via adaptive learning rates and task-specific fine-tuning.

\section{Conclusions}
We introduce a novel data generation paradigm to overcome data scarcity and limited diversity. By synthesizing 62,000 QA pairs and 73,000 object descriptions from ScanNet, ScanQA, and ScanRefer, our approach enriches training data and enhances performance in vision-and-language tasks. Furthermore, our model effectively encodes 3D point clouds with attention mechanisms and object detection cues, reducing ambiguities and establishing a robust framework for future 3D vision and language integration.

\section{Acknowledgements}

This work is supported by National Key Research and Development Program of China (2024YFE0203100). It is also supported by the Beijing Natural Science Foundation (No. JQ23014), and in part by the National Natural Science Foundation of China (No. 62271074), Open Project Program of the State Key Laboratory of Virtual Reality Technology and Systems, Beihang University (No. VRLAB2025B03).



\vspace{12pt}


\bibliographystyle{IEEEtran}
\bibliography{IEEEabrv,References}

\newpage

\twocolumn[{%
 \centering
 \LARGE [Supplementary Material] \\ 3D-MoRe: Unified Modal-Contextual Reasoning for Embodied Question Answering\\[1.5em]
}]

\section*{APPENDIX}
\subsection{Data Distribution}
\begin{figure}[htbp]
\centering
\includegraphics[width=1\linewidth]{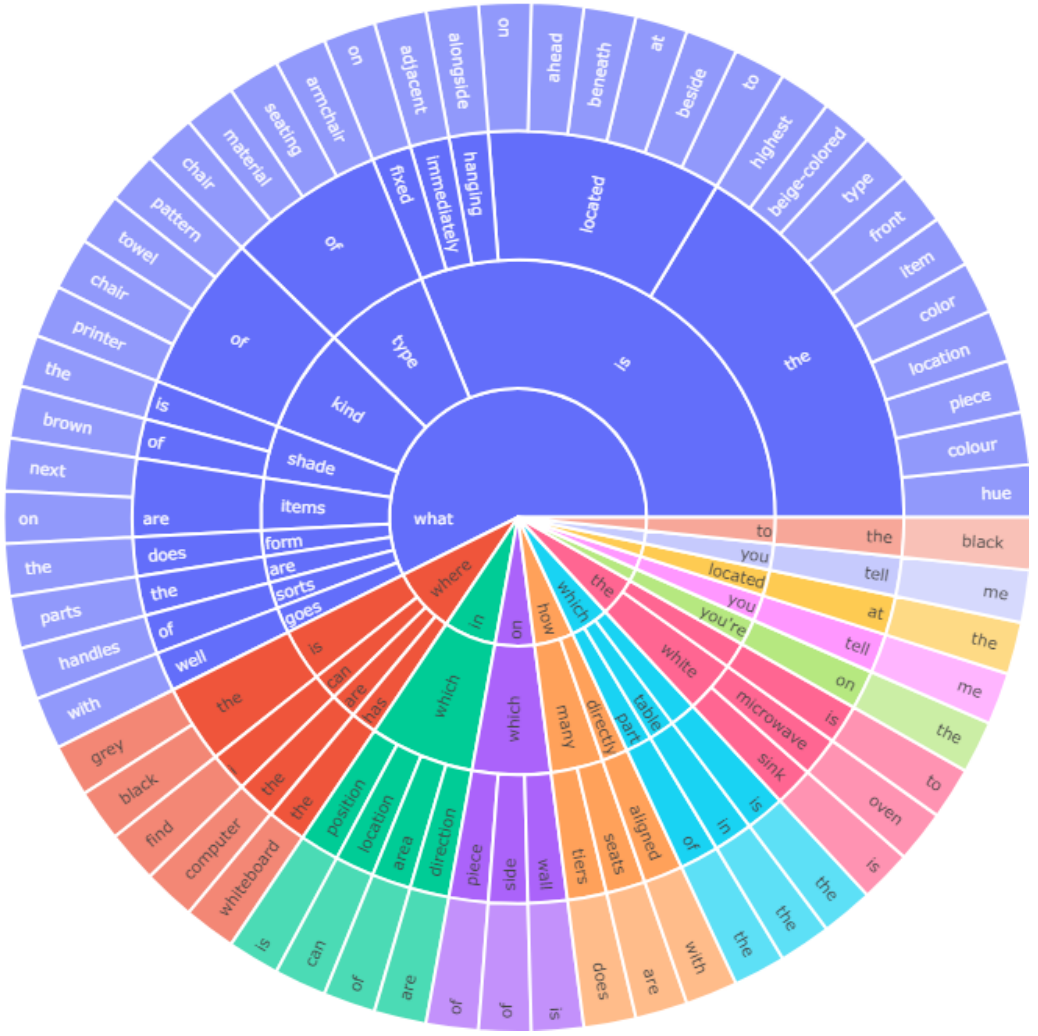}
\vspace{-0.2cm}
\caption{we analyzed the distribution of the first four types of questions based on the expanded dataset. The results indicate that these questions cover several aspects: Firstly, there are questions concerning local objects, such as "What color is it?" and "What type of thing is it?"; Secondly, there are questions concerning the global context, such as "How many are there?" and "Are there more than how many tables?"; Thirdly, there are questions involving complex relationships between multiple objects, such as "What is to the right of the trash can?" and "What is in the middle?"; Finally, there are questions concerning directions or positions, such as "Where is it?" and "Which way is it facing?".}
\label{question_distribution}
\end{figure}

\begin{figure}[htbp]
\centering
\includegraphics[width=1\linewidth]{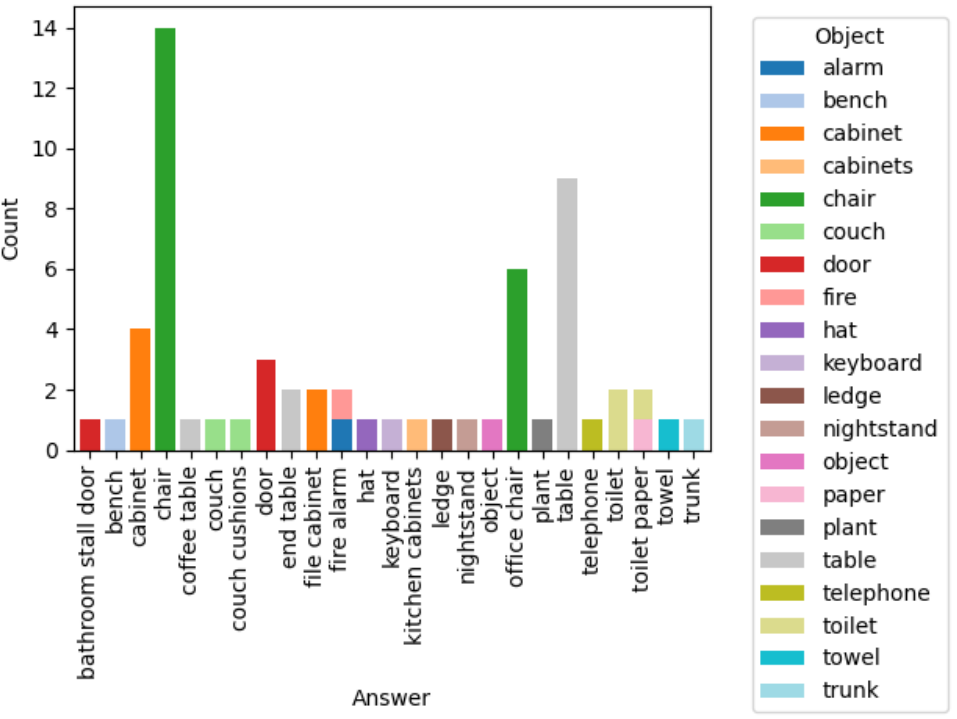}
\vspace{-0.2cm}
\caption{We analyzed the top 20 words that appeared most frequently in the responses based on an expanded dataset, showing the distribution of answers to different types of questions. The X-axis of the chart represents the answers, the Y-axis represents the count, and different colors represent different words.".}
\label{answer_distribution}
\end{figure}

Figure.\ref{question_distribution} illustrates the analysis of the first four question types based on an expanded dataset, revealing four distinct aspects: inquiries about local objects (e.g., "What color is it?"), questions regarding the global context (e.g., "How many are there?"), complex relational questions (e.g., "What is to the right of the trash can?"), and those addressing directions or positions (e.g., "Where is it?"). Meanwhile, Figure \ref{answer_distribution} presents the distribution of answers by highlighting the top 20 most frequently occurring words in responses, where the X-axis indicates the answers, the Y-axis reflects their counts, and different colors distinguish the words.

\begin{table*}[ht]
\caption{\textbf{Evaluation of Scaling Data Paradigm on Model Performance}  Bold: This indicator outperforms the other tasks being compared, with CIDEr serving as the primary evaluation metric.}
\centering
\renewcommand{\arraystretch}{1.8}
\begin{tabular}{|c|c|ccccccc|}
\hline
\multirow{2}{*}{\textbf{Dataset}} & \multicolumn{7}{c|}{\textbf{Evaluation Metrics}} \\ 
\cline{2-8}
 & \textbf{BLEU-1} & \textbf{BLEU-2} & \textbf{BLEU-3} & \textbf{BLEU-4} & \textbf{CIDEr} & \textbf{ROUGE-L} & \textbf{METEOR} \\
\hline
3D-LLM \cite{hong20233d}(BLIP2-flant5) & \textbf{39.30} & \textbf{25.20} & \textbf{18.40} & 12.00 & 69.40 & \textbf{35.70} & 14.50 \\
Custom Dataset (50 epochs) & 35.37 & 22.07 & 15.38 & 10.71 & 69.63 & 35.37 & 14.31 \\
Custom Dataset (100 epochs) & 37.93 & 25.10 & 18.14 & \textbf{13.13} & \textbf{72.68} & 35.42 & \textbf{15.14} \\
\hline
\end{tabular}
\label{more_evaluations}
\end{table*}
\begin{figure}[ht]
\centering
\includegraphics[width=1\linewidth]{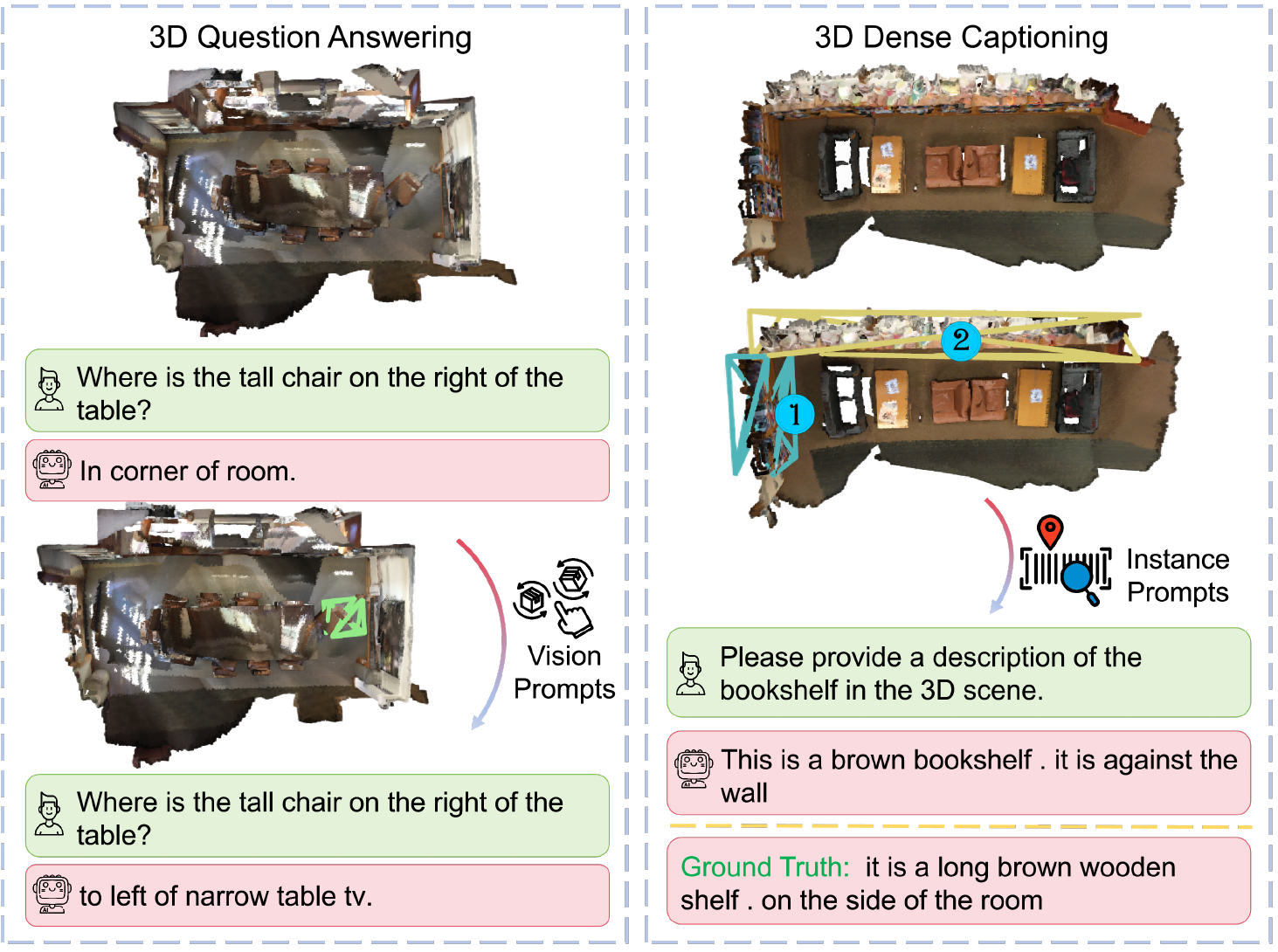}
\vspace{-0.2cm}
\caption{Vision and Instance Prompts improve object localization and distinction, enhancing model accuracy in 3D captioning and question answering.}
\label{qualitative_results}
\vspace{-0.5cm}
\end{figure}
\subsection{More Qualitative Results}
As shown in Figure.\ref{qualitative_results}, In the 3D Dense Captioning and 3D Question Answering tasks, Vision Prompts and Instance Prompts play crucial roles in guiding the model's understanding and generation of outputs. Vision Prompts help in the spatial localization of objects by framing the specific region or object in question, ensuring that the model focuses on the correct target within the 3D environment. This is particularly important in complex scenes where multiple objects may overlap or clutter the view. On the other hand, Instance Prompts assign unique identifiers to the objects being described, facilitating the model's ability to distinguish between multiple instances of similar objects. This instance-level annotation helps improve the accuracy of the model's responses by allowing it to produce more structured and specific descriptions, crucial for both QA and dense captioning tasks in rich 3D environments.
\begin{table}[ht]
\caption{\textbf{Evaluation of Scaling Data Paradigm on Different Backbone Versions} 3D-MoRe: Using the Scaling Data Paradigm we proposed. Bold: This indicator outperforms the other tasks being compared, with CIDEr serving as the primary evaluation metric. Note: The model used in this table is the LL3DA\cite{chen2024ll3da} with different Backbone Versions.}
\renewcommand{\arraystretch}{1.8}
\setlength{\tabcolsep}{3pt} 
\centering
\begin{tabular}{|c|cccc|}
\hline
\textbf{Dataset} & \multicolumn{4}{c|}{\textbf{Evaluation Metrics}} \\ 
\cline{2-5}
& \textbf{BLEU-4} & \textbf{CIDEr} & \textbf{ROUGE-L} & \textbf{METEOR} \\
\hline
opt-1.3b & 13.53 & 76.79 & 37.31 & 15.88 \\
opt-1.3b(3D-MoRe) & \textbf{14.17} & 78.94 & 37.89 & \textbf{16.07} \\
\hline
Qwen1.5-7B & 13.61 & 78.78 & 38.09 & 15.95 \\
Qwen1.5-7B(3D-MoRe) & 13.98 & \textbf{79.22} & \textbf{38.41} & 16.00 \\
\hline
\end{tabular}
\label{backbone_versions}
\vspace{-8pt}
\end{table}
\subsection{More Evaluations}
In this section, we evaluate the effectiveness of the Scaling Data Paradigm by applying the newly extended dataset to multiple models, particularly focusing on the 3DLLM's performance in 3D Question Answering tasks(As shown in Table.\ref{more_evaluations}). The goal is to investigate whether scaling the dataset improves model performance across key evaluation metrics, including BLEU, CIDEr, ROUGE-L, and METEOR. As the table shows, the model trained on 100 epochs with the extended dataset demonstrates notable improvements, particularly in BLEU-4 (+1.13) and CIDEr (+3.28) scores, indicating enhanced capability in generating coherent and contextually relevant answers in complex 3D scenes. This suggests that the Scaling Data Paradigm is effective in improving model performance as the data volume increases, leading to better understanding and reasoning within 3D environments.

\subsection{More backbone version}
As shown in Table.\ref{backbone_versions}, We evaluate the impact of the extended dataset generated using the Scaling Data Paradigm on different backbone versions of the LL3DA model: `facebook/opt-1.3b`\cite{zhang2022opt} and `Qwen/Qwen1.5-7B`\cite{yang2024qwen2}. As seen in the table, applying the expanded dataset consistently improves the performance across both backbones, particularly in CIDEr and BLEU-4 metrics. For instance, in the case of `facebook/opt-1.3b`, the BLEU-4 score increases from 13.53 to 14.17, and CIDEr improves from 76.79 to 78.94. Similarly, the Qwen backbone shows an increase in BLEU-4 from 13.61 to 13.98, and CIDEr from 78.78 to 79.22. These improvements demonstrate the effectiveness of the Scaling Data Paradigm in enhancing the model's ability to comprehend and generate accurate responses in the 3D environment, with both larger and smaller backbone versions benefiting from increased data diversity.

\vspace{12pt}

\end{document}